\documentclass[11pt,a4paper]{article}
\usepackage{multirow}
\usepackage{booktabs}

\usepackage{times}
\usepackage{latexsym}
\usepackage{hhline}
\usepackage{xspace}
\usepackage{graphicx}
\usepackage{subcaption}

\newcommand{\example}[1]{``\textit{#1}"}

\newcommand{\component}[1]{{\small $\langle$\tt #1$\rangle$\xspace}}
\newcommand{\comp}[1]{{\scriptsize $\langle$\tt #1$\rangle$\xspace}}

\usepackage{tabularx} % in the preamble
\newcommand{\squishlist}{
 \begin{list}{$\bullet$}
  { \setlength{\itemsep}{0pt}
     \setlength{\parsep}{1pt}
     \setlength{\topsep}{1pt}
     \setlength{\partopsep}{0pt}
     \setlength{\leftmargin}{1em}
     \setlength{\labelwidth}{1em}
     \setlength{\labelsep}{0.5em} } }
 \newcommand{\squishend}{\end{list}}
\newcommand{\eat}[1]{}

\usepackage{latexsym}
\usepackage{microtype}
\usepackage[hyperref]{emnlp2020}

\aclfinalcopy % Uncomment this line for the final submission
\title{Learning Structured Representations of Entity Names using Active Learning and Weak Supervision}

\author{Kun Qian,\hspace{5pt} Poornima Chozhiyath Raman,\hspace{5pt}Yunyao Li,\hspace{5pt}Lucian Popa\\
  IBM Research\\
  \texttt{qian.kun@ibm.com}, \texttt{\{pchozhi, yunyaoli, lpopa\}@us.ibm.com} \\
  }

\date{}

\begin{document}
\maketitle

\begin{abstract}
\looseness=-1 
%Entity names usually have structured representation that is useful for many entity-related tasks such as entity normalization and variant generation. Learning the structured representation of entity names in low-resource settings without context and external knowledge bases is challenging. In this paper, we present a novel learning framework that combines active learning and weak supervision to solve this problem, and we experimentally show that our method can learn high-quality BERT-CRF models in low-resource settings. A video demo of a system that implements this framework is included in supplementary materials.
Structured representations of entity names are useful for many entity-related tasks such as entity normalization and variant generation. Learning the implicit structured representations of entity names without context and external knowledge 
is particularly challenging. In this paper, we present a novel learning framework that combines active learning and weak supervision to solve this problem. Our experimental evaluation show that this framework enables the learning of high-quality models from merely a dozen or so labeled examples. 
\end{abstract}

\section{Introduction}
Entity normalization and variant generation are fundamental for a variety of other tasks such as semantic search and relation extraction \cite{bhutani2018exploiting, Arasu:2009}. Given an entity name $E$, the goal of entity normalization is to convert $E$ to a canonical form (e.g., ``\textit{Jordan, Michael}" $\rightarrow$ ``\textit{Michael Jordan}"), while the goal of entity variant generation is to convert $E$ to a set of different textual representations that refer to the same entity as \textit{E} (e.g., ``\textit{Michael Jordan}" $\rightarrow$ \{``\textit{Jordan, Michael}", ``\textit{MJ}", ``\textit{M. Jordan}", $\dots$\}). 

\looseness=-1 Typically, entity normalization and variant generation are done by first performing entity linking \cite{moro2014entity, zhao2019neural, li2017cnn}, i.e., matching entity names appearing in some context (e.g., free text) to named entities in curated knowledge bases (KBs), then use the canonical form or variations (of the linked entities) residing in the KBs to complete the tasks. Unfortunately, in some scenarios, such as search \cite{thompson-dozier-1997-name}, entity names are not surrounded by context. 
Furthermore, for specialized domain-specific applications, there may not be a knowledge base to govern the names of the relevant entities. Thus, entity linking is not always applicable.
In this paper, 
we take the view 
%the problem differently, in particular, we argue 
that entity normalization and variant generation can be done without contextual information or 
external KBs if we understand the internal structures of entity names.

% Fundamental to the success of entity linking is the availability of the contextual information (i.e., free texts where the entity names appear) and ontological information (i.e., properties and neighbors of named entities in KBs).  

%  and there are no external KBs that we can use as master datasets to match input entity names. \todo{One may argue that DBpedia and Wikipedia is a good proxy. It may be useful to talk about related work taking this view here.}
% For example, when searching \example{General Electric Company}, we need to also consider variations like \example{GE Co.}, \example{G.E.}, \example{General Electric}, etc. without relying on any contextual information and external KBs. 
% Performing entity normalization and variant generation in a contextless fashion is extremely challenging because we have only the surface forms of entity names.

% Several attempts have been made to parse the structured representation of entity names.
As observed in \cite{campos2015entity, bhutani2018exploiting, Arasu:2009,katiyar-cardie-2018-nested, finkel2009nested}, entity names often have 
%structured representation 
implicit structures 
that can be exploited to solve entity normalization and variant generation. Table \ref{tab:samples} shows how we can manipulate such structured representations of entity names to generate different variations without help from context or external knowledge.

% For example, if we know that \example{Michael} is \component{first} and \example{Jordan} is \component{last} in \example{Michael Jordan}, we can generate two variations: (1) \example{M. Jordan} (by converting its \component{first} to initial \example{M.}) and (2) \example{Jordan, Michael} (by switching the order of \component{first} and \component{last} and adding a comma right after \component{last}).

\begin{table*}[ht]
\centering
\scriptsize
\begin{tabular}{|c|l|l|l|}
\hline
\textbf{Mention}                          & \textbf{Structured Representation}           & \multicolumn{1}{c|}{\textbf{Manipulation}} & \multicolumn{1}{c|}{\textbf{Variations}} \\ \hline
\multirow{3}{*}{\textit{Michael Jordan}}           & \multirow{3}{*}{``\textit{Michael}"\comp{first} ``\textit{Jordan}"\comp{last}}  & \comp{last},\comp{first}                                & \textit{Jordan, Michael}                          \\ \cline{3-4} 
                                          &                              & \textit{createInitial}(\comp{first}) \comp{last}                       & \textit{M Jordan}                                 \\ \cline{3-4} 
                                          &                              & \textit{createInitial}(\comp{first}) \textit{createInitial}(\comp{last})            & \textit{MJ}                                       \\ \hhline{|=|=|=|=|}
\multirow{2}{*}{\textit{General Electric Company}} & \multirow{2}{*}{``\textit{General Electric}"\comp{name} ``\textit{Company}"\comp{suffix}} & {\textit{createInitial}}(\comp{name}) \textit{drop}(\comp{suffix})               & \textit{GE}                                       \\ \cline{3-4} 
                                          &                              & \textit{createInitial}(\comp{name}) \textit{abbreviate}(\comp{suffix})         & \textit{GE Co.}                                   \\ \hline
\end{tabular}
\caption{Normalization \& variant generation by manipulating structured representation of entity names}
\label{tab:samples}
\end{table*}

Declarative frameworks are proposed in \cite{Arasu:2009, campos2015entity} to allow 
%high-skill 
developers to manually specify rules that parse 
entity names into 
%the 
a structured representation.
%of enity names.
To avoid such low-level manual effort, \cite{katiyar-cardie-2018-nested, finkel2009nested} used fully supervised methods for identifying nested entities embedded in flat named entities. Unfortunately, labeled data are rarely available to leverage these methods in the real-world. 
To mitigate the need for training data, \cite{bhutani2018exploiting, qian2018lustre} proposed an active learning
%-based 
system, LUSTRE, to semi-automatically learn rules for mapping entity names to their structured representations. 
By 
%making use of 
using regex-based extractors and a list of comprehensive dictionaries that capture crucial domain vocabularies, LUSTRE can generate rules that achieve SoTA results. However, for more complex and realistic scenarios, dictionaries may not be available and regex-based extractors alone are not expressive enough. Moreover, as shown in Section~\ref{sec:experiments}, LUSTRE cannot handle long entities such as machine logs.

In this paper, we present a 
%learning 
framework that learns high-quality BERT-CRF models for parsing entity names into  
structured representations 
%of entity names 
in low-resource settings, namely, when no labeled data is available. The proposed framework is essentially an active learning-based approach that learns from human interactions. We believe that comprehensible user interfaces are essential for active learning-based approaches, especially for labeling tasks that require non-trivial human labels (e.g., sequence labels in our approach). Therefore, we developed a system named PARTNER \cite{qian2020partner} that implements this framework. We designed the interface of PARTNER similar to that of LUSTRE, but we also made major modifications so that it is more user friendly. 
Interested readers can find a video demo of PARTNER at \url{http://ibm.biz/PARTNER}.
Our main contributions include:
% \todo{Low-resource setting can mean different things. It would be helpful to clearly describe what you mean here.}

% {
% \squishlist
%     \item We developed a full-fledged system built upon an effective framework to learn high-quality BERT-CRF models for parsing structured representation of entity names without contextual information  (seee the video demo).
%     \item To minimize human effort, our framework combines active learning and weak supervision, which were usually applied in isolation.
    
%     \item Both the datasets and the system will be made publicly available. 
% \squishend
% }

{
\squishlist
    \item 
    %We propose 
    A hybrid framework combining active learning and weak supervision to effectively learn BERT-CRF-based models with low human effort. 
    \item 
    %We developed 
    A full-fledged system, with intuitive UI, that implements the framework.
    \item Comprehensive experimental results showing that the framework learns high-quality models 
    from merely a dozen or so labeled examples. 
\squishend
}

\medskip
\textbf{Related work.}
Our problem is related to both flat and nested named entity recognition (NER). However, as discussed in \cite{finkel2009nested}, NER focuses on identifying the outermost flat entities and completely ignores their internal structured representations. \cite{katiyar-cardie-2018-nested, ju-etal-2018-neural, finkel2009nested, dinarelli-rosset-2012-tree} identify nested entities within some context using fully supervised methods that require large amounts of labeled data, whereas our goal is to learn from very few labels (e.g., $< 15$) in a contextless fashion. 
Active learning \cite{settles2009active} and weak supervision have been widely adopted for solving many entity-centric problems, such as entity resolution \cite{kasai-etal-2019-low, qian2019systemer, qian2017active, ertutorial-cikm19}, NER \cite{lison2020named, shen2018deep,he2017unified, nadeau2007semi}, and entity linking \cite{chen2011el}. While the power of the combination of the two techniques has been demonstrated in other domains (e.g., computer vision \cite{brust2020active}), to the best of our knowledge, the two approaches are usually applied in isolation in prior entity-related work.

Recently, data programming approaches (e.g., \cite{Snorkel, safranchik2020weakly}) use labeling functions/rules to generate weak labels to train machine learning models in low-resource scenarios. Data programming approaches like Snorkel usually assume that labeling functions are manually provided by users, indicating that their target users must have programming skills in order to provide such labeling functions. In contrast, our goal is to minimize both human effort (i.e., minimize labeling requests) and lower human skills (no programming skills are needed).

\section{Methodology}
\begin{figure}[ht]
    \centering
    \includegraphics[width=0.48\textwidth]{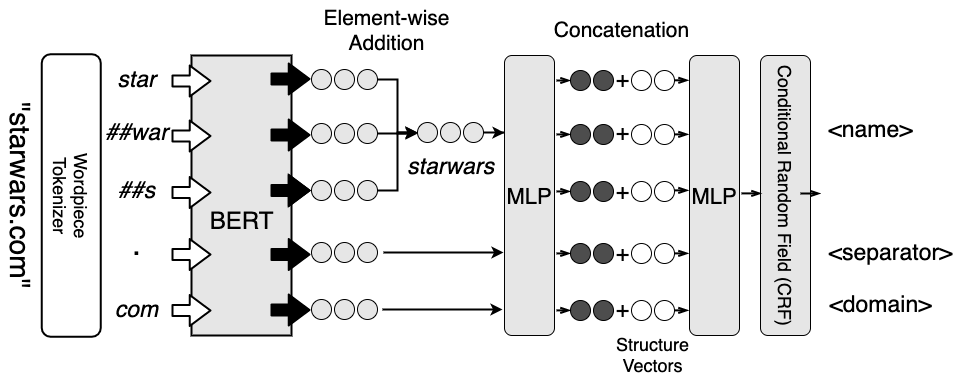}
    \caption{BERT-CRF based model}
    \label{fig:architecture}
\end{figure}

\looseness=-1 Given a set $E=\{\mathcal{E}_1, \dots, \mathcal{E}_m\}$ of isolated entity mentions (name strings) of a particular type, where $\mathcal{E}_i$ is a sequence $\mathcal{E}_i=(t_i^1,\dots,t_i^n)$ of tokens. Assume that the input set $E$ of entity names contain a set $C=\{\mathcal{C}_1,\dots,\mathcal{C}_k\}$ of semantic components (i.e., labels such as \component{first}, \component{middle} in person names). Our goal is to learn a labeling model $\mathcal{M}: \mathcal{E}_i=(t_i^1,\dots,t_i^n) \rightarrow (y_1,\dots,y_n)$, where $y_k\in C$. 
The labeling model $\mathcal{M}$ is a BERT-CRF based model (see Fig. \ref{fig:architecture}) with several key modifications, which we elaborate next. 

\medskip
\noindent\textbf{Tokenization \& vectorization}. 
An input entity name is tokenized with BERT's wordpiece tokenizer, which may result in sub-words for out-vocabulary tokens, e.g., \example{starwars} $\rightarrow$ \{\example{star}, \example{\#\#war}, \example{\#\#s}\}. In this case, we combine these sub-words' embeddings (from BERT) into one vector using element-wise addition (see Fig. \ref{fig:architecture}). 
We then feed the sequence of token embeddings to a multi-layer perceptron (MLP), the goal of which is to condense the BERT embeddings to smaller embeddings (e.g., 50), so that they are somewhat comparable to the size of the \textit{structure vectors} (to be discussed next), which are crucial for our active learning and weak supervision approach. It is not hard to see that the pre-trained BERT model can be replaced with any other seq2seq models with pretrained static word embeddings such as BiLSTM + fastText \cite{bojanowski2016enriching}.

\medskip
\noindent \textbf{Structure vectors}. We predefined a set of boolean predicates where each of them verifies whether or not a token satisfies a specific syntactic pattern. In our experiments, we defined a list of 15 predicates, which can be easily extended, as shown below:

{
\small
\tt
    \noindent hasAllCapsTokens()\\
    hasAllLowerTokens()\\
    hasAllAlphbeticalToken()\\
    hasPunctuationOnly() \\
    isAlphanumToken()\\
    containsNumber()\\
    containsPunctuation()\\
    isFirstLetterCapitalized()\\
    isTwoDigitNumber()\\
    isFourDigitNumber()\\
    isSingleDigitNumber()\\
    isInteger()\\
    isNumericToken()\\
    appearAtBegining()\\
    appearAtEnd()\\
}

% {
%     {
%     \small
%     \squishlist
%     \item \predicate{isAllCaps(t)} - if $t$ has all capitalized letters.
%     \item \predicate{isFirstCaps(t)} - if the first letter of $t$ is capitalized.
%     \item \predicate{endsWithDot(t)} - if $t$ ends with a dot.
%     \item \predicate{hasNbr(t)} - if $t$ contains any character that is a digit.
%     % \item \predicate{isNonword(t)} - if $t$ is not alphanumeric.
%     \squishend
%     }
% }

\noindent Each token is then converted to a boolean vector using the predefined boolean predicates, and is concatenated with the corresponding condensed token embedding emitted from the first MLP (see Fig \ref{fig:architecture}). Intuitively, condensed token embeddings can capture semantic information and structure vectors can capture structural information. 
% We will describe how we use structure vectors for active learning and weak supervision shortly.

\medskip
\noindent\textbf{CRF layer}. Each of the concatenated vector are fed to another MLP, which condense them into a vector of size $|C|$ (i.e., the number of label classes). Finally, the final CRF layer uses viterbi algorithm to find out the most likely sequence of labels using the emission vectors (i.e., embeddings from the last MLP layer) and learned transition matrix. 
% Together with the sequence of labels, CRF will also produce a probability score $S$, we further divide it by the number of the input tokens to get the normalized probability score $\bar{S}$, which is used in our active learning strategy.

% \section{Active Learning and Weak Supervision}
% Our learning framework is essentially a human-in-the-loop framework such that the model learning algorithm updates the BERT-CRF model based on human interactions. To make it truly usable in low-resource settings, we carefully combines active learning and weak supervision (to lower human effort) and surface the technology with an intuitive user interface that provides end-to-end support for the whole learning process. A short video demo of the implemented system is included in the supplementary materials, which visually illustrates the human-machine co-cooperate experiences for learning the labeling model as well as performing normalization and variant generation based on the learned model. 

\subsection{Weak Supervision with Structure Vectors}
Recall that each token is associated with a binary structure vector that carries its ``structure" information. Consider the following company names:
% \begin{center}
% {
% \small
% ``Apple Inc."\quad ``Microsoft Corp."\quad ``Coca Cola Co."
% }
% \end{center}

\smallskip
{
{\footnotesize
\bf
\squishlist
\item ``Apple Inc." = \{\example{Apple}, \example{Inc.}\}
\item ``Microsoft Corp." = \{\example{Microsoft}, \example{Corp.}\}
\item ``Coca Cola Co." = \{\example{Coca}, \example{Cola}, \example{Co.}\}
\squishend
}
}
\smallskip

\looseness=-1 \noindent Although textually dissimilar, they are structurally identical. Concretely, {\example{Apple}}, {\example{Microsoft}}, {\example{Coca}}, and {\example{Cola}} all contain only alphabetical letters with the first one capitalized; Tokens {\example{Inc.}}, {\example{Corp.}}, and {\example{Co.}} all are alphabetical letters with first letter capitalized, and they all end with a dot. Therefore, {\example{Apple Inc.}} and {\example{Microsoft Corp.}} have the same sequence of structure vectors. Moreover, for consecutive tokens with identical structure vectors, we combine them into one and hence {\example{Coca Cola}}
shares the same structure vectors with the other two. 
% In this way, all the three are structurally identical. 
Therefore, if one of the three is labeled as \component{name}\component{suffix}, we can apply the same sequence of labels to the other two examples as weak labels without actual human annotation.

To some extend, the structure vector-based weak supervision approach adopted in our framework is similar to the labeling functions/rules adopted in data programming approaches (e.g., \cite{Snorkel}). In our framework, predefined boolean predicates can be viewed as token-level labeling functions, which are later automatically combined as entity-level labeling functions (together with condensed BERT embeddings) used by the second MLP in our architecture (see Figure \ref{fig:architecture}). Moreover, in our framework, the labeling functions are transparent to the user, thus no programming skills are needed. 

\subsection{Active Sampling Strategy}
% \vspace{-2mm}
% \begin{figure}[ht]
%     \vspace{-2mm}
%     \centering
%     \includegraphics[width=0.45\textwidth]{img/workflow.png}
%     % \vspace{}
%     \vspace{-2mm}
%     \caption{Active learning flow}
%     \label{fig:interaction}
%     \vspace{-4mm}
% \end{figure}

The model learning process has multiple iterations, where each starts with requesting the user to label the entity with highest \textit{informative score} (to be defined shortly). Based on the user labeled entity, a set $k$ of other entities with identical sequence of structure vectors will be automatically labeled and used for incrementally updating the model being learned. Then, unlabeled entities are annotated by the refined model and ranked according to the probability scores produced by the CRF layer. Subsequently, both top-$p$ high-confidence and bottom-$q$ low-confidence machine-label entities are sent to the user for verification (i,e, correct or incorrect).
% Optionally, the top-$p$ high-confidence labeled entities can also be directly added to our labeled dataset as weak labels. The bottom-$q$ low-confidence labeled entities are sent to the user for verification (i,e, correct or incorrect). 
We also update the unlabeled entity set by removing user labeled entities and weakly labeled entities. We repeat the process until either user's labeling budget is completed or most (e.g, $\ge$ 90\%) of the low-confidence labeled entities are correct. 

\medskip
\noindent\textbf{Informative Score.} The informativeness of an entity is measured according to its \textit{representativeness} and \textit{uncertanty}. 
Let $\mathcal{S}(\mathcal{E}_i)$ denote the sequence of structure vectors of entity $\mathcal{E}_i$, then we define the representativeness of $\mathcal{E}_i$ with respect to the current set $E^{u}$ of unlabeled entity as follows:

\smallskip
{
\small
~~~~~~~~~~$\textup{Rep}(\mathcal{E}_i)=|~\{\mathcal{E}_k~|~ \mathcal{S}(\mathcal{E}_k)=\mathcal{S}(\mathcal{E}_i), \forall \mathcal{E}_k\in E^{u}\}~|$
}
\smallskip

% \[
% \small
% \textup{Rep}(\mathcal{E}_i)=|~\{\mathcal{E}_k~|~ \mathcal{S}(\mathcal{E}_i)=\mathcal{S}(\mathcal{E}_k), \forall \mathcal{E}_k\in E\}~|.
% \]

\noindent Intuitively, the representativeness of an entity is the total number of entities in the unlabeled data that have the same sequence of structure vectors. The uncertainty score of an entity $\mathcal{E}_i$ is defined as:

\smallskip
{
%\vspace{1mm}
\small
\quad\quad\quad\quad\quad
$\displaystyle \textup{Uncertain}(\mathcal{E}_i)=\frac{1}{Pr(\mathcal{M}(\mathcal{E}_i)) / |\mathcal{E}_i|}$
}
\smallskip

% \[
% \small
% \textup{Uncertain}(\mathcal{E}_i)= \frac{1}{Pr(\mathcal{M}(\mathcal{E}_i))},
% \]

\noindent where $Pr(\mathcal{M}(\mathcal{E}_i))$ is the probability score of the most likely sequence of labels for $\mathcal{E}_i$ produced by the final CRF layer, and $|\mathcal{E}_i|$ is the number of tokens in $\mathcal{E}$ (divided by this term to normalize the probability score wrt the length of the entities). Then, the informative score of an entity $\mathcal{E}_i$ is:

\smallskip
%\vspace{1mm}
{
\small
~~~~\quad\quad\quad$\textup{Info}(\mathcal{E}_i) = \textup{Rep}(\mathcal{E}_i) \times\textup{Uncertain}(\mathcal{E}_i).$
}

% \[
% \small
% \textup{Info}(\mathcal{E}_i) = \textup{Rep}(\mathcal{E}_i) \times \textup{Uncertain}(\mathcal{E}_i).
% \]
\smallskip
\noindent Thus, informative examples are the ones that are structurally highly representative and for which the current model is highly uncertain.

\section{Experimental Evaluation}
\label{sec:experiments}
\begin{table}[ht]
\centering
\scriptsize
\begin{tabular}{|c|c|l|}
\hline
\multicolumn{1}{|c|}{\textbf{Type}} & \multicolumn{1}{c|}{\textbf{\# entities}} & \multicolumn{1}{|c|}{\textbf{Components}} \\ \hline
\multirow{2}{*}{\textbf{PER}}       & \multirow{2}{*}{1302}                     & \comp{title}\comp{first}\comp{middle}\comp{last}               \\
                                    &                                           & \comp{suffix}\comp{degree}                \\ \hline
\textbf{ORG}                        & 2209                                      & \comp{corename}\comp{type}\comp{suffix}\comp{location}                     \\ \hline
\textbf{DATE}                       & 1190                                      & \comp{Year}\comp{MonthOfYear}\comp{Day}                     \\ \hline
\multirow{2}{*}{\textbf{LOG}}       & \multirow{2}{*}{1323}                     & \comp{host}\comp{time}\comp{filename}\comp{operation}                \\
                                    &                                           & \comp{requesttype}\comp{errormsg}\comp{remainder}                \\ \hline
\end{tabular}
\caption{Statistics of datasets}
\label{tab:dataset}
\end{table}

\begin{table*}[ht]
\centering
\scriptsize
\begin{tabular}{|c|p{0.45\textwidth}|p{0.45\textwidth}|}
\hline
\multicolumn{1}{|l|}{\textbf{Type}} & \textbf{Sample Mentions}      & \textbf{Structured Representation} \\ \hline
\multirow{3}{*}{\textbf{PER}}       & \textit{Prof Liat Sossove}                                                                                & \comp{title}\comp{first}\comp{last}                         \\ \cline{2-3} 
                           & \textit{Hagop Youssoufia, B.S.}                                                                           & \comp{first}\comp{last},\comp{degree}                         \\ \cline{2-3} 
                           & \textit{ElmeDxsna Adzemovi Sr.}                                                                           & \comp{first}\comp{last}\comp{suffix}                          \\ \hhline{|=|=|=|}
\multirow{3}{*}{\textbf{ORG}}       & \textit{SONY CORP.  }                                                                                     & \comp{corename}\comp{suffix}                        \\ \cline{2-3} 
                           & \textit{JONES APPAREL GROUP INC}                                                                          & \comp{corename}\comp{corename}\comp{type}\comp{suffix}                        \\ \cline{2-3} 
                           & \textit{STAPLES, INC.}                                                                                    & \comp{corename},\comp{suffix}  \\ \hhline{|=|=|=|}
\multirow{3}{*}{\textbf{DATE}}      & \textit{February 2, 2019}                                                                                 & \comp{MonthOfYear}\comp{Day},\comp{Year}                         \\ \cline{2-3} 
                           & \textit{6/13/2012}                                                                                        & \comp{MonthOfYear}/\comp{Day}/\comp{Year}                        \\ \cline{2-3} 
                           & \textit{1st day of April 2019}                                                                            & \comp{Day}\comp{tok}\{2\}\comp{MonthOfYear}\comp{Year}                         \\ \hhline{|=|=|=|}
\multirow{2}{*}{\textbf{LOG}}       & \textit{719+1: Tue Aug 22 08:26:41 1995 (/wow/wow-mbos.gif): Sent binary: GET /wow/wow-mbos.gif HTTP/1.0} & \comp{host}\comp{timestamp}(\comp{filename})\comp{operation}: \comp{requestType} \comp{filename} \comp{remainder}                         \\ \hline
\end{tabular}
\caption{Sample entity mentions and their expected structured representations from each dataset}
\label{tab:sample_entities}
\end{table*}

\looseness=-1 We implemented the system with Pytorch \cite{pytorch_nips} and pytorch-transformer \cite{Wolf2019HuggingFacesTS}. Four different entity types were considered (see Table \ref{tab:dataset}). Sample mentions of each entity type and corresponding expected structured representations are given in Table \ref{tab:sample_entities}. Two baselines: (1) \textbf{CRF-AW} \cite{CRFsuite}: linear-chain conditional random field using our structure vectors as features (2) \textbf{LUSTRE}: the prior SoTA active learning system for learning structured representations of entity names \cite{bhutani2018exploiting}.
More details about datasets, the implementation of the system, best-performing hyperparameter settings, and evaluation metrics can be found at \url{https://github.com/System-T/PARTNER}.

% \textit{Hyperparameters setting, datasets, training strategy, dictionaries for LUSTRE are included in supplementary materials}.

{
\begin{figure*}[ht!]
    \begin{subfigure}[t]{0.25\textwidth}
        \centering
        \includegraphics[width=\textwidth]{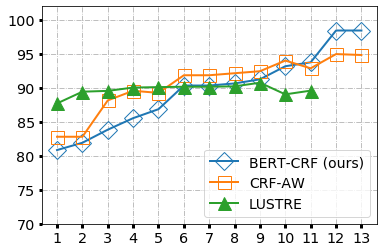}
        \caption{Entity-level for PER}
    \end{subfigure}%
    \begin{subfigure}[t]{0.25\textwidth}
        \centering
        \includegraphics[width=\textwidth]{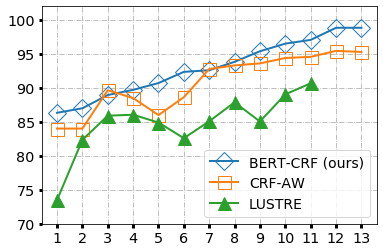}
        \caption{Token-level for PER}
    \end{subfigure}
    \begin{subfigure}[t]{0.25\textwidth}
        \centering
        \includegraphics[width=\textwidth]{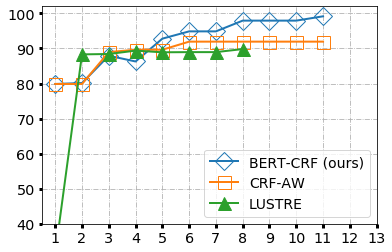}
        \caption{Entity-level for ORG}
    \end{subfigure}%
    \begin{subfigure}[t]{0.25\textwidth}
        \centering
        \includegraphics[width=\textwidth]{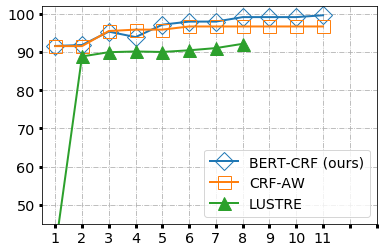}
        \caption{Token-level for ORG}
    \end{subfigure}
    \\
    \begin{subfigure}[t]{0.25\textwidth}
        \centering
        \includegraphics[width=\textwidth]{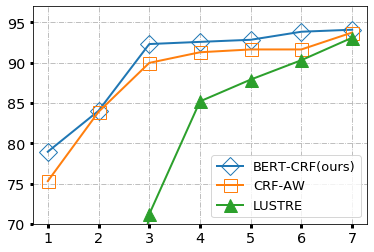}
        \caption{Entity-level for DATE}
    \end{subfigure}%
    \begin{subfigure}[t]{0.25\textwidth}
        \centering
        \includegraphics[width=\textwidth]{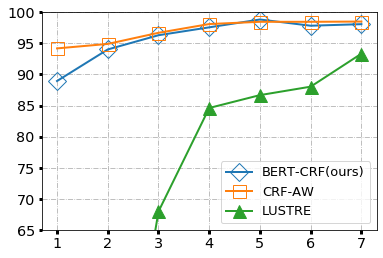}
        \caption{Token-level for DATE}
    \end{subfigure}
    \begin{subfigure}[t]{0.25\textwidth}
        \centering
        \includegraphics[width=\textwidth]{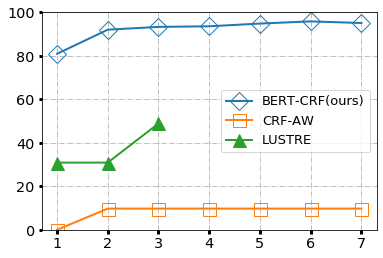}
        \caption{Entity-level for LOG}
    \end{subfigure}%
    \begin{subfigure}[t]{0.25\textwidth}
        \centering
        \includegraphics[width=\textwidth]{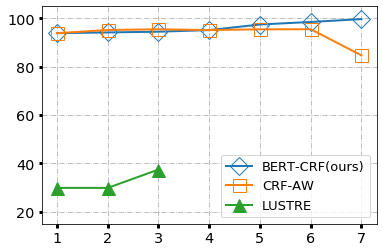}
        \caption{Token-level for LOG}
    \end{subfigure}
    \caption{F1-scores at entity and token levels (X-axis: \# iterations (i.e., \# actual user labels); Y-axis: F1-scores)}
    \label{fig:results}
\end{figure*}
}

\begin{table}[ht]
\centering
\scriptsize
\begin{tabular}{cl|l|l|l|}
\cline{3-5}
                                            &              & \textbf{Ours}  & \textbf{CRF-AW} & \textbf{LUSTRE} \\ \hline
\multicolumn{1}{|c|}{\multirow{7}{*}{\textbf{PER}}}  & \comp{first}    & \textbf{0.989} & 0.958           & 0.97            \\ \cline{2-5} 
\multicolumn{1}{|c|}{}                      & \comp{middle}   & \textbf{0.88}  & 0.835           & 0.81            \\ \cline{2-5} 
\multicolumn{1}{|c|}{}                      & \comp{last}     & \textbf{0.975} & 0.967           & 0.92            \\ \cline{2-5} 
\multicolumn{1}{|c|}{}                      & \comp{suffix}       & 0.761          & 0.70            & \textbf{0.821}  \\ \cline{2-5} 
\multicolumn{1}{|c|}{}                      & \comp{nickname}     & \textbf{0.989} & 0.80            & 0.23            \\ \cline{2-5} 
\multicolumn{1}{|c|}{}                      & \comp{degree}       & \textbf{0.991} & 0.972           & 0.787           \\ \cline{2-5} 
\multicolumn{1}{|c|}{}                      & \comp{title}        & \textbf{1}     & 0.765           & 0.652           \\ \hline
\multicolumn{1}{|c|}{\multirow{4}{*}{\textbf{ORG}}}  & \comp{corename}     & \textbf{0.996} & 0.963           & 0.971           \\ \cline{2-5} 
\multicolumn{1}{|c|}{}                      & \comp{suffix}       & \textbf{0.969} & 0.792           & 0.923           \\ \cline{2-5} 
\multicolumn{1}{|c|}{}                      & \comp{location}     & \textbf{0.976} & 0.976           & 0.78            \\ \cline{2-5} 
\multicolumn{1}{|c|}{}                      & \comp{type}         & \textbf{0.991} & 0.0             & 0.243           \\ \hline
\multicolumn{1}{|c|}{\multirow{3}{*}{\textbf{DATE}}} & \comp{MonthOfYear}  & \textbf{0.988} & 0.964           & 0.942           \\ \cline{2-5} 
\multicolumn{1}{|c|}{}                      & \comp{Day}          & \textbf{0.941} & 0.939           & 0.916           \\ \cline{2-5} 
\multicolumn{1}{|c|}{}                      & \comp{Year}         & \textbf{0.957} & \textbf{0.957}  & 0.944           \\ \hline
\multicolumn{1}{|c|}{\multirow{7}{*}{\textbf{LOG}}}  & \comp{Host}         & \textbf{1}     & 0.994           & 0.486           \\ \cline{2-5} 
\multicolumn{1}{|c|}{}                      & \comp{time}         & \textbf{1}     & \textbf{1}      & 0.489           \\ \cline{2-5} 
\multicolumn{1}{|c|}{}                      & \comp{filename}     & \textbf{0.998} & 0.756           & 0.560           \\ \cline{2-5} 
\multicolumn{1}{|c|}{}                      & \comp{operation}    & \textbf{0.993} & 0.739           & 0.520           \\ \cline{2-5} 
\multicolumn{1}{|c|}{}                      & \comp{remainder}    & \textbf{0.994} & 0.823           & 0.529           \\ \cline{2-5} 
\multicolumn{1}{|c|}{}                      & \comp{errorMessage} & \textbf{0.991} & 0.239           & 0               \\ \cline{2-5} 
\multicolumn{1}{|c|}{}                      & \comp{requestType}  & \textbf{1}     & 0.603           & 0.527           \\ \hline
\end{tabular}
\caption{Final F1-scores for different components}
\label{tab:components}
\end{table}

\looseness=-1 For our system, we ask the user to label the example with highest informative score (e.g., label \example{Michael} as \component{first} and \example{Jordan} as \component{last}) in each active learning iteration. Then, $k=50$ (a hyperparameter) structurally similar examples will be automatically labeled. In each iteration, 51 new labeled examples (or less, since there may not be 50 structurally similar examples)  will be collected and used to incrementally refine the model. Since CRF-AW is fully supervised, we give it the sets of labels we iteratively accumulated during the active learning of our model. Hence, CRF-AW is not vanilla CRF models, they are enhanced by our structure vectors, our active learning, and our weak supervision strategies. 
% \footnote{We got an implementation of LUSTRE from the authors.}. 
% Since CRF is fully supervised, we gave it the labeled datasets created by our BERT-CRF model in different iterations. More details about the experiment settings can be found in supplementary materials.

\smallskip
\noindent\textbf{Metrics}. We report F1-scores at \textit{entity-level}, \textit{token-level}, and \textit{component-level}.
Entity-level measures how well the model correctly labels individual entities (all tokens of an entity must be correctly labeled). For token and component-level results, we apply models to make predictions for each token in the given set of test entities. Each token prediction is credited as correct if it matches the true label. The difference between the token and component-level evaluation is that the former accumulates the credits over all tokens regardless the actual classes they belong to, whereas the latter evaluation accumulates the credits with respect to the actual classes.

\smallskip
\noindent\textbf{Results.} 
% The F1-scores achieved in each active learning iteration of these methods are reported in Figure \ref{fig:results}.
\looseness=-1 Figure \ref{fig:results} reports entity-level and token-level results for all methods at different iterations. As can be seen, our approach consistently outperforms the baselines, requiring only 7 to 13 actual user annotations per task. 
Moreover, as the active learning goes, the F1-score curves of our method in all tasks increase monotonically, showing stable performance. Supported by the weak labels obtained by our active sampling strategy, CRF-AW gives the suboptimal results (except for the entity-level performance for LOG), but there are still noticeable gaps between CRF-AW and ours, indicating that pre-trained BERT still plays an essential role. 
LUSTRE fails to match our performance except for DATE. This finding is not surprising: since LUSTRE learns highly precise rules from user labeled examples, its recall is largely determined by its sampling strategy, which is less effective to find a variety of structurally diverse examples. 
Since our method always make a prediction, recall is trivially 100\%, but the overall precision for entity-level and token-level is relatively low initially. Then, whether or not our sampling strategy can keep finding the most informative examples to ``complete" the training set is crucial for enhancing the overall precision. 
The monotonically increasing F1 curves indicate confirm the effectiveness of our method.

% The experiments also show that our structure-vector-based sampling strategy is highly effective to find informative examples.
%
% We observed interesting results from the LOG scenario, where entities are long texts with complex structures. CRF-AW performs well at token level but bad at entity level. This indicates that it can correctly label most tokens in an entity, but makes minor mistakes leading to entity-level errors. LUSTRE outperforms CRF since it learns very precise rules, but LUSTRE terminated after three iterations (in multiple runs) because it runs out of memory when it tries to learn a rule consisting of more than 70 regex primitives to capture a long log message. Regarding the component-level results, as shown in Table \ref{tab:components}, our methods significantly outperform other baselines, which is not a surprise given that our method gives the best token-level results. CRF-AW again gives the suboptimal results with the help from our active learning and weak supervision.
%

The LOG dataset consists of entities with long text and complex structures. For this dataset, CRF-AW performs well at token level but bad at entity level, indicating that it can correctly label most tokens in an entity name, but makes minor mistakes leading to entity-level errors. LUSTRE outperforms CRF-AW since it learns very precise rules. However, LUSTRE terminated quickly as it runs out of memory when trying to learn a rule with over 70 regex primitives to capture a long log message. 

Regarding the component-level results, as shown in Table \ref{tab:components}, our methods significantly outperform other baselines, which is not a surprise given that our method gives the best token-level results. 
% CRF-AW again gives the suboptimal results with the help from our active learning and weak supervision.

% results

% For our models and CRF-AW, the recall is trivially 100\% for entity-level and token-level evaluations because they always make a prediction for an entity or token. It is not true for LUSTRE, which learns very specific and nearly 100\% precise rules from examples it sees, but its recall is affected by its active sampling strategy that did not find a variety of structurally representative examples quickly.

% \textbf{Efficiency.} Experiments were run on a machine with 8-core CPUs. For BiLSTM-CRF models, one iteration (including both labeling and model refinement) takes about 3 minutes, showing reasonable interactive experience. BERT-CRF models require more than 20 minutes per iteration. Both can be improved by using GPU machines. 

\section{Concluding Remarks}
\looseness=-1 We proposed a framework for learning structured representation of entity names under low-resource settings. In particular, we focus on a challenging scenario, where entity names are given as textual strings without context. Experiments show the efficacy of our approach. One immediate future work is to generate explanations for model predictions using structured vector. 
% Future directions include using character-level models and providing explanations for model predictions using structure vectors.
% Experiments over real-world datasets show that our approach can learn high-quality models with a small number of human labels. 

\section*{Acknowledgments}
We would like to thank the anonymous reviewers for their critical reading and constructive suggestions, which improve and clarify this paper.

\bibliography{emnlp2020}
\bibliographystyle{acl_natbib}

\appendix 

\section{Appendix}
In the appendix, we provide more details about the experimental evaluation presented in the main paper. Other useful materials are also included in the github repository at  \url{https://github.com/System-T/PARTNER}.

\subsection{Datasets}
We studied four different datasets, which will be made publicly available in the github repo mentioned earlier after we finish the open source approval process. 

In our experiments, each dataset contains a training set and a held-out test set, which we will release.  The datasets are manually annotated, and we adopted the 70\%-30\% splitting convention. Concretely, we first label all examples, and then 70\% of them became the ``unlabeled" data that is going to be provided to our system, and the rest 30\% became a held-out test set. 

\begin{table}[ht]
\scriptsize
\centering
\begin{tabular}{|llc|}
\hline
\multicolumn{3}{|c|}{\textbf{Input Representation}}                       \\
% BiLSTM-CRF word embedding size          &  & 300                 \\
BERT-CRF word embeddings size           &  & 768                 \\
Input dropout rate                      &  & 0                   \\
% \multicolumn{3}{|c|}{\textbf{Token-level BiLSTM (for BiLSTM-CRF only)}}   \\
% LSTM hidden size                        &  & 50                  \\
% LSTM output size                        &  & 50                  \\
% Number of LSTM layers                   &  & 1                   \\
% Final output (forward, backward)        &  & concat              \\
\multicolumn{3}{|c|}{\textbf{Structure Vectors}}                          \\
size of predicates                      &  & 15                  \\
\multicolumn{3}{|c|}{\textbf{Multilayer Perceptron Layer}}                          \\
MLP-1 (after BERT)   (input, output) &  & (768,50)            \\
MLP-2 (after Concatenation)    (input, output) &  & (50, \# labels) \\
Activation function                     &  & relu                \\ \hline
\multicolumn{3}{|c|}{\textbf{Training}}                                   \\
Optimization                            &  & SGD                 \\
\# epochs                                &  & 30                  \\
Learning rate                           &  & 0.01                \\
Learning rate decay                     &  & $1\times10^{-4}$    \\
Loss function                           &  & Negative log likelihood \\\hline
\multicolumn{3}{|c|}{\textbf{Active Learning \& Weak Supervision Parameters}}        \\
\# structurally similar examples         &  & 50                  \\
\# high-confidence examples              &  & 15                  \\
\# low-confidence examples               &  & 15                  \\ \hline
\end{tabular}
\caption{Architecture, hyperparameters, and training}
\label{tab:hyper}
\end{table}

\subsection{Model Implementation}
Table \ref{tab:hyper} lists the hyperparameters of the best-performing model that we reported in the main paper. As mentioned earlier, we used huggineFace pytorch-transformer \cite{Wolf2019HuggingFacesTS} to implement our BERT-CRF model. In particular, we used the pretrained {\small \tt BertModel}, and we obtained the tokenizer and pretrained weights using the {\tt \small bert-base-cased} configuration. 

The Bert embedding size (i.e., 768) is predefined by the Bert model, and the size of predicates for structural vectors are predefined by us. The input-output dimension size of MLP-1 (bounds are determined by the embedding size of Bert and number of labels) and learning rate are determined using random sampling (using the DATE dataset).

\subsection{Evaluation Metrics}
We used three metrics in the main paper: \textit{entity-level}, \textit{token-level}, \textit{component-level}. Here we give a concrete example for computing the three metrics. Consider a test set consisting of a single DATE entity string: \textbf{June 3rd, 2020}.
Assume that we have three models: $m_1$, $m_2$, and $m_3$. The predictions of the three models over the single test date entity is shown in Table \ref{tab:metric}, and the F1-scores of these models at entity-level, token-level, and component-level are shown Table \ref{tab:results}.

\begin{table}[ht]
\small
\centering
\begin{tabular}{@{}c|c|c|c|@{}}
\cmidrule(l){2-4} & \textbf{\color{red} June} & \textbf{\color{red} 3rd} & \textbf{\color{red} 2020} \\ \midrule
\multicolumn{1}{|c|}{\textbf{Ground truth}} & \textbf{month}         & \textbf{day}          & \textbf{year}          \\ \midrule
\multicolumn{1}{|c|}{{$m_1$'s predictions}}           & month         & day          & year          \\ \midrule
\multicolumn{1}{|c|}{{$m_2$}'s predictions}           & month         & month        & year          \\ \midrule
\multicolumn{1}{|c|}{{$m_3$'s predictions}}           & none          & none         & none          \\ \bottomrule
\end{tabular}
\caption{Predictions of three dummy models (none means the model does not make a prediction)}
\label{tab:metric}
\end{table}

\begin{table}[ht]
\centering
\scriptsize
\begin{tabular}{@{}c|l|l|l|l|@{}}
\cmidrule(l){2-5}
\multicolumn{1}{l|}{}                              & \textbf{Entity-level}                                                                              & \textbf{Token-level}                                                                                & \multicolumn{2}{l|}{\textbf{Component-level}}                                           \\ \midrule
\multicolumn{1}{|c|}{\multirow{3}{*}{\textbf{$m_1$}}} & \multirow{3}{*}{1.0}                                                                               & \multirow{3}{*}{1.0}                                                                                & month & 1.0                                                                             \\ \cmidrule(l){4-5} 
\multicolumn{1}{|c|}{}                             &                                                                                                    &                                                                                                     & day   & 1.0                                                                             \\ \cmidrule(l){4-5} 
\multicolumn{1}{|c|}{}                             &                                                                                                    &                                                                                                     & year  & 1.0                                                                             \\ \midrule
\multicolumn{1}{|c|}{\multirow{3}{*}{\textbf{$m_2$}}} & \multirow{3}{*}{\begin{tabular}[c]{@{}l@{}}0.0\\ (precision = 0)\\ (recall    = 1.0)\end{tabular}} & \multirow{3}{*}{\begin{tabular}[c]{@{}l@{}}0.80\\ (precision = 0.67)\\ (recall = 1.0)\end{tabular}} & month & \begin{tabular}[c]{@{}l@{}}0.67\\ (precision=0.5)\\ (recall = 1.0)\end{tabular} \\ \cmidrule(l){4-5} 
\multicolumn{1}{|c|}{}                             &                                                                                                    &                                                                                                     & day   & \begin{tabular}[c]{@{}l@{}}0\\ (precision=1.0)\\ (recall = 0.0)\end{tabular}    \\ \cmidrule(l){4-5} 
\multicolumn{1}{|c|}{}                             &                                                                                                    &                                                                                                     & year  & 1.0                                                                             \\ \midrule
\multicolumn{1}{|l|}{\multirow{3}{*}{\textbf{$m_3$}}} & \multirow{3}{*}{0.0}                                                                               & \multirow{3}{*}{0.0}                                                                                & month & 0.0                                                                             \\ \cmidrule(l){4-5} 
\multicolumn{1}{|l|}{}                             &                                                                                                    &                                                                                                     & day   & 0.0                                                                             \\ \cmidrule(l){4-5} 
\multicolumn{1}{|l|}{}                             &                                                                                                    &                                                                                                     & year  & 0.0                                                                             \\ \bottomrule
\end{tabular}
\caption{F1-scores of the three dummy models at entity-level, token-level, and component-level}
\label{tab:results}
\end{table}

Given that $m_1$ correctly predict all the tokens (thus the entire entity), it is easy to see that it's F1-scores are all 1.0's. Similar argument can also show that $m_3$'s F1-scores are all 0's. For $m_2$, since it does not correctly label all the three tokens of the entity string, so the entity-level is 0. However, it does correctly labeled two tokens (i.e., ``June" and ``3rd") out of the three tokens, so the precision is $0.67=2/3$. Moreover, since $m_2$ makes predictions for all the three tokens, thus the recall is trivially 100\%, which means it's token-level F1 is 0.80. In fact, DL-based approaches such as ours that takes a sequence of tokens as input, will trivially achieve 100\% recall. For component-level evaluation, $m_2$ predicted two tokens as \textbf{month}, where only one of them is correct, so the precision for month component is 50\%. Since $m_2$ identified all the true \textbf{month} tokens, the recall is 100\%. For \textbf{day} component, $m_2$ does not predict any token as \textbf{day}, but the second token ``3rd" has true label \textbf{day}, so the recall is 0\%, which leads to an F1-score of 0. We do not discuss its \textbf{year} component performance because it is obvious 100\%.

\subsection{Training}
Unlike typical deep learning-based supervised learning approaches, where there are a lot of labeled examples, we have limited training data in each active learning iteration. It does not make sense to split the limited number of labeled examples (e.g., about 30) into a training set and a development set, and use the development set to choose the best-performing model. First, the number of examples is too small to make the splitting meaningful. Second, we could potentially perform k-fold cross validation, but that would require much more time, which makes the user experience bad (i.e., the user has to wait for a long time before the training is done). 

To make the system as interactive as possible, we used a simple heuristic, that is, we simply train the model with a fixed number of epoch (with random shuffling after each epoch), in our experiments, we set the number to 30 epochs. However, we would terminate the training early if the difference between the total loss of two consecutive epochs is less than a certain threshold, which is set to 10$^{-3}$ in our experiments. The main intuition is that we want to let the model somewhat overfit the training data as they are considered to be ``informative" based on our active learning strategy, but we do need to avoid ``extreme" overfitting.

\subsection{Environment and Runtime}
We have run our experiments both on a CPU machine (Apple Macbook Pro 2019 model) and on a GPU machine (with 1 Tesla V100 GPU). Recall that our framework is active learning-based, where each active learning iteration contains three steps: (1) user labeling, (2) model updating, and (3) user feedback. The most time-consuming part is the model updating phase. For the experiments reported in this paper, depending on the sizes of labeled data, the model updating phase takes 10 to ~70 seconds with the GPU machine, and 30 seconds to 10 mins with the CPU machine.

\end{document}